\pdfoutput=1

\documentclass[11pt]{article}

\usepackage[final]{acl}
\usepackage{float}
\usepackage{stfloats}
\usepackage{graphicx}
\usepackage{multirow}
\usepackage{booktabs}
\usepackage{arydshln}
\usepackage{amsmath,amsfonts,amssymb}
\usepackage{tablefootnote}
\usepackage{bbm}
\usepackage{times}
\usepackage{latexsym}

\usepackage[T1]{fontenc}

\usepackage[utf8]{inputenc}

\usepackage{microtype}

\usepackage{inconsolata}

\newcommand{\ctslogo}{\raisebox{3.4pt}{\includegraphics[scale=0.0105]{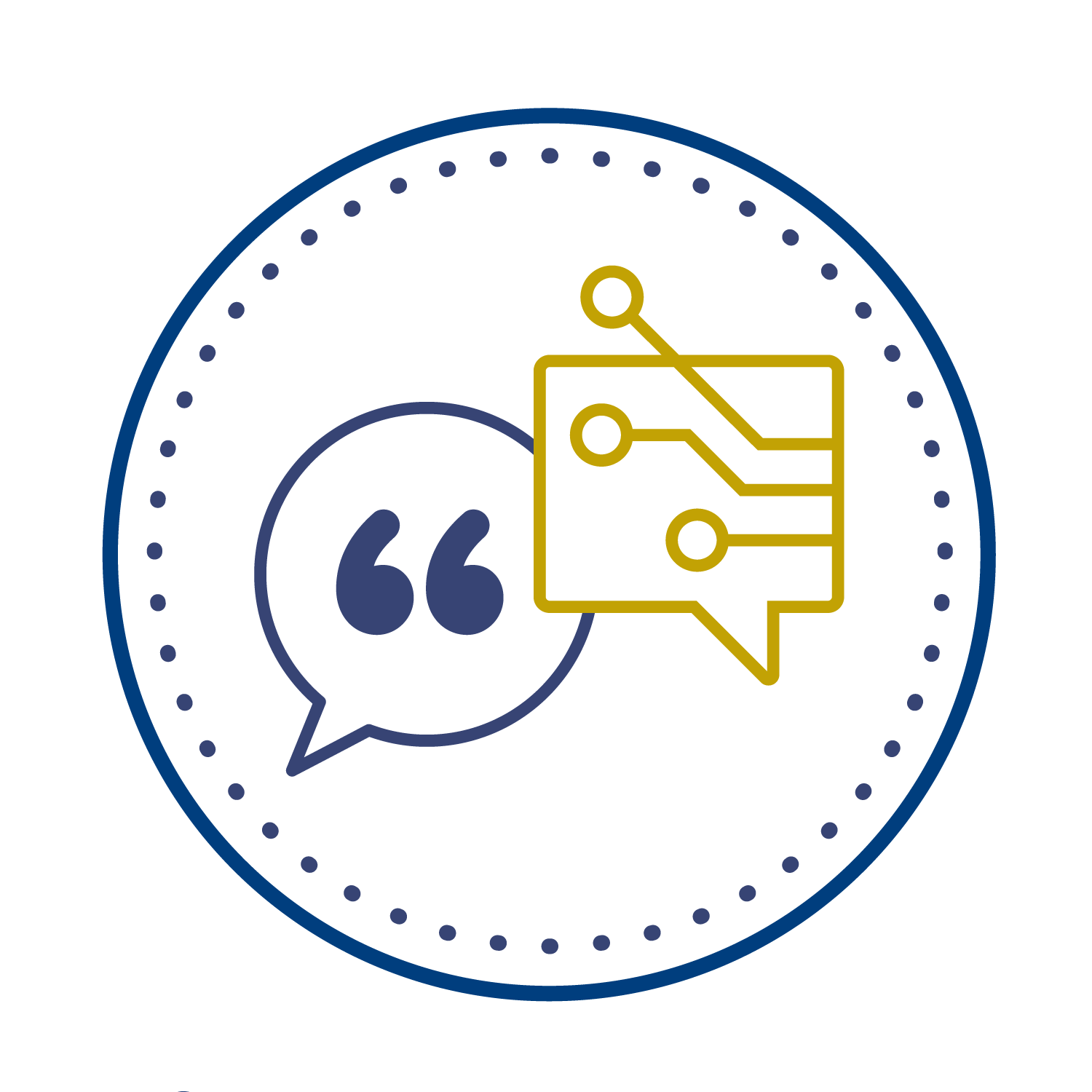}}}
\newcommand{\PAIlogo}{\raisebox{3.4pt}{\includegraphics[scale=0.080]{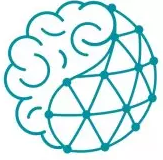}}}

\title{A Multi-task Learning Framework for Evaluating Machine Translation \\ of Emotion-loaded User-generated Content}

\author{Shenbin Qian\ctslogo, Constantin Orăsan\ctslogo, Diptesh Kanojia\PAIlogo  and Félix do Carmo\ctslogo \\
\ctslogo Centre for Translation Studies and 
\PAIlogo Institute for People-Centred AI, \\[.2em]
University of Surrey, United Kingdom \\[.13em]
\{s.qian, c.orasan, d.kanojia, f.docarmo\}@surrey.ac.uk}

\begin{document}
\setlength{\abovedisplayskip}{3pt}
\setlength{\belowdisplayskip}{6pt}

\maketitle
\begin{abstract}
Machine translation (MT) of user-generated content (UGC) poses unique challenges, including handling slang, emotion, and literary devices like irony and sarcasm. Evaluating the quality of these translations is challenging as current metrics do not focus on these ubiquitous features of UGC. To address this issue, we utilize an existing emotion-related dataset that includes emotion labels and human-annotated translation errors based on Multi-dimensional Quality Metrics. We extend it with sentence-level evaluation scores and word-level labels, leading to a dataset suitable for sentence- and word-level translation evaluation and emotion classification, in a multi-task setting. We propose a new architecture to perform these tasks concurrently, with a novel combined loss function, which integrates different loss heuristics, like the Nash and Aligned losses. Our evaluation compares existing fine-tuning and multi-task learning approaches, assessing generalization with ablative experiments over multiple datasets. Our approach achieves state-of-the-art performance and we present a comprehensive analysis for MT evaluation of UGC.
\end{abstract}

\section{Introduction}

\begin{figure*}[h]
  \centering
  \includegraphics[width=0.9\textwidth]{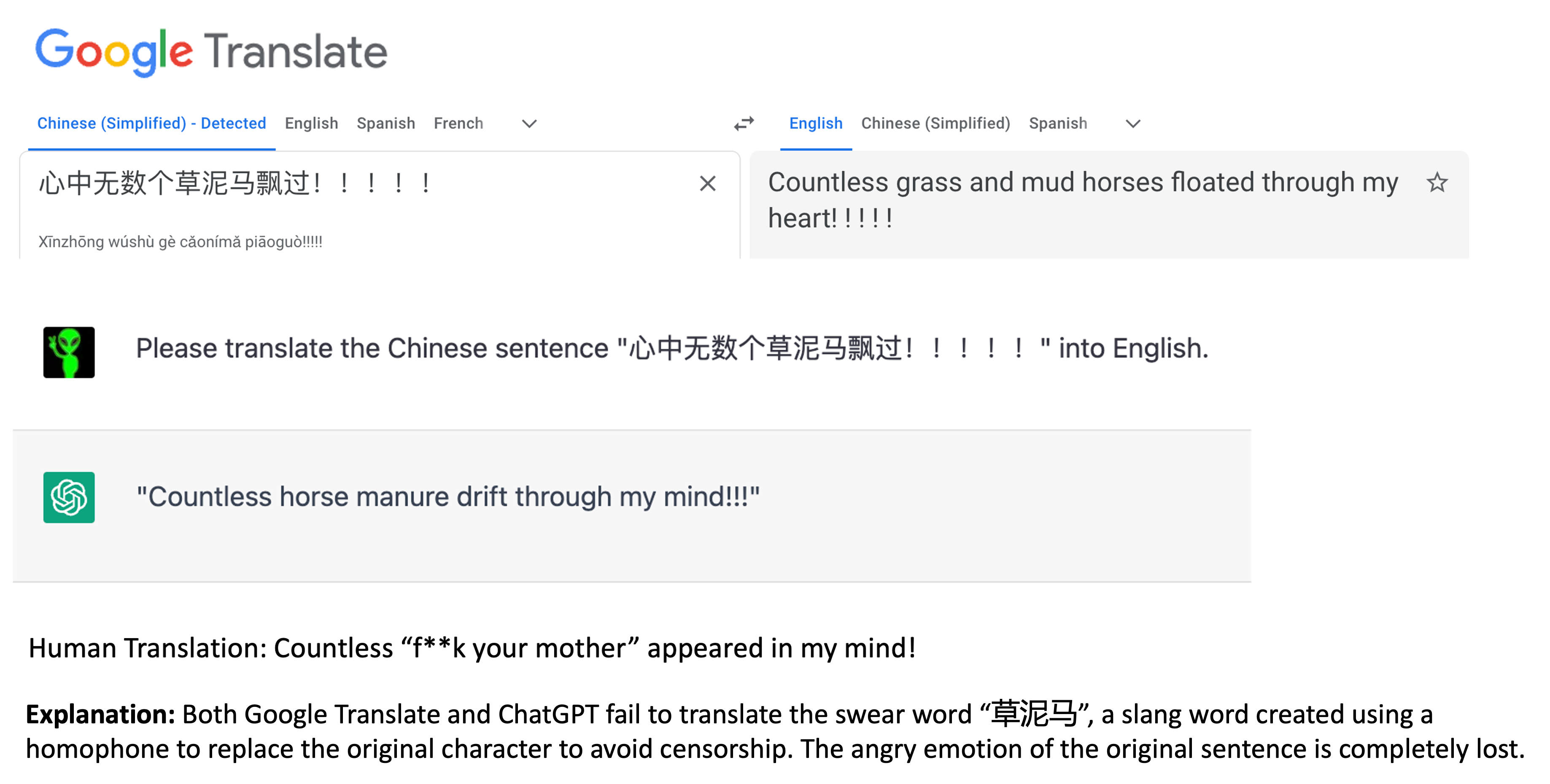}
  \caption{Example of translations from Google Translate and ChatGPT}
  \label{fig.example1}
\end{figure*}

Machine translation (MT) has advanced rapidly in recent years, leading to claims it has achieved human parity in Chinese-English news translation~\citep{Hassan2018}. Recent advent of large language models (LLMs) has determined researchers to repeat claims of human parity more often~\citep{Wang2021-lh}. However, automatically translating user-generated content (UGC) with expressions that contain emotions, like tweets, reveals novel challenges for MT systems~\citep{saadany-etal-2023-analysing}. Figure~\ref{fig.example1} shows the output of Google Translate (GT) and ChatGPT when we translated some Chinese UGC with emotional slang using them\footnote{GPT-3.5 at ``https://chat.openai.com/'' in April, 2024}. As can be seen from the example, both outputs need to be improved significantly to be considered usable. Similar problems were noticed with other MT engines, indicating that it is imperative to evaluate MT quality with metrics that take emotion preservation into account. 

Using human judgements/input to evaluate MT quality is expensive in terms of both time and money~\citep{Dorr2011, Lai2020}. Quality estimation (QE), which predicts MT quality in the absence of human references, can serve as a cost-effective alternative to approximate human evaluation based on metrics like Multi-dimensional Quality Metrics (MQM), an error-based human evaluation scheme for MT quality~\citep{Lommel2014}. A widely-used approach in QE involves fine-tuning a multilingual pre-trained language model (PTLM) using human evaluation data~\citep{blain-etal-2023-findings}. This fine-tuned model can predict scores for entire MT sentences or labels for individual words, indicating whether each word is correctly translated or not. This encompasses two common QE tasks: sentence-level QE and word-level QE.

To assess MT quality of emotion-loaded UGC, it is crucial to evaluate the overall quality of emotion preservation after translation (sentence-level QE), and how well emotion words are translated (word-level QE). To achieve this, we leverage an existing emotion-related dataset that includes emotion labels and MQM-based human-evaluated translation errors. We extend it with sentence-level QE scores and word-level labels, resulting in a dataset extension. This extended dataset is suitable for both sentence- and word-level QE, and emotion classification. For joint training of these tasks, we employ multi-task learning (MTL), anticipating improved performance for all tasks due to their inherent correlation with emotionally charged content. We further introduce a new architecture with a novel combined loss function that integrates different loss heuristics, enabling the concurrent training of these tasks and optimizing their overall performance. We compare our MTL approach with existing fine-tuning and MTL methods. Our proposed approach achieves new state-of-the-art results on the emotion-related QE dataset and a standard QE dataset. Our contributions can be summarized as follows: 

\begin{itemize}
\itemsep 0mm

\item \textit{Extending an emotion-related QE dataset} with 1) QE scores at sentence level and 2) labels indicating emotion-related translation quality at word level. 
\item A new architecture with a \textit{novel combined loss function}, integrating different loss heuristics for multi-task learning\footnote{Our code and the extended dataset for MTL are available at \url{https://github.com/shenbinqian/MTL4QE.}}.
\item Evaluation of the proposed MTL approach on multiple QE datasets including ablative experiments on combinations of QE and emotion classification tasks, \textit{improving performance over existing fine-tuning and MTL methods}. 

\end{itemize}

Section~\ref{lit} discusses existing work for QE and MTL while Section~\ref{data} introduces the datasets we use for this study. Our approach, baselines and experimental setup are described in Section~\ref{method}, and Section~\ref{results} discusses the results obtained on multiple datasets. Section~\ref{conclusion} concludes our study and outlines future directions. Section~\ref{limitations} points out limitations and ethical considerations. Relevant mathematical equations and loss algorithms are explained in Appendix~\ref{sec:appendix}.

\section{Related Work} \label{lit}

We discuss related work in supervised QE in $\S$~\ref{review-qe}. Studies on MTL and its application to QE are reviewed in $\S$~\ref{review-mtl}.

\subsection{Quality Estimation} \label{review-qe}

Though prompting with LLMs is increasingly applied to the field of quality evaluation~\citep{kocmi-federmann-2023-large,kocmi-federmann-2023-gemba,fernandes-etal-2023-devil}, supervised fine-tuning of multilingual PTLMs on human evaluation data based on metrics such as translation edit rate~\citep{snover-etal-2006-study}, direct assessment~\citep{graham-etal-2013-continuous} and MQM, remains as state-of-the-art QE methods~\citep{kocmi-federmann-2023-large}. TransQuest~\citep{ranasinghe-etal-2020-transquest} and COMET~\citep{rei-etal-2020-comet, stewart-etal-2020-comet, rei-etal-2022-cometkiwi, Guerreiro2023} are two popular frameworks used for sentence-level QE. TransQuest utilizes XLM-RoBERTa~\citep{conneau-etal-2020-unsupervised} as the backbone, concatenating the source and target sentences using [CLS] (start) and [SEP] (separator) tokens. In MonoTransQuest, an architecture within TransQuest, only the embeddings of the [CLS] token are used for prediction. In SiameseTransQuest, a variant of TransQuest architecture, a twin XLM-RoBERTa network computed the mean of all token embeddings for the source and target. This mean is then used to calculate the cosine similarity as the final QE score. Unlike TransQuest, COMET was initially proposed for reference-based evaluation until 2022, when COMETKIWI \citep{rei-etal-2022-cometkiwi} was introduced to support reference-less evaluation. Similar to MonoTransQuest, it concatenates the source and target, and inputs them into the encoder. All hidden states are then fed into a scalar mix module~\citep{peters-etal-2018-deep} that learns a weighted sum, producing a new sequence of aggregated hidden states. The output of the [CLS] token is then used for the prediction of sentence-level QE scores. 

For word-level QE, OpenKiwi~\citep{kepler-etal-2019-openkiwi} was proposed to support both sentence- and word-level QE with various neural network architectures. MicroTransQuest~\citep{ranasinghe-etal-2021-exploratory}, utilizing outputs of all input tokens of an XLM-RoBERTa model based on the MonoTransQuest architecture, was proposed only for word-level QE under multilingual settings. 

Because of their successes in the QE shared tasks in the Conference on Machine Translation (WMT) in recent years~\citep{specia-etal-2020-findings-wmt, specia-etal-2021-findings, zerva-etal-2022-findings}, TransQuest and COMET are selected as our baseline fine-tuning frameworks for sentence-level QE, and MicroTransQuest for word-level QE.

\subsection{Multi-task Learning} \label{review-mtl}

Multi-task learning addresses multiple related tasks concurrently by training them simultaneously with a shared representation~\citep{Caruana1997}. While this approach reduces the training cost compared to training separate models~\citep{Baxter2000}, early methods led to performance degradation when compared to single-task models~\citep{Standley2020}. Recent efforts have introduced various methods to address this problem and enhance the MTL performance.

\citet{Liu2019-dwa} proposed dynamic weight averaging that could learn task-specific feature-level attention. They used a shared network that contains features of all tasks and a soft-attention module for each specific task without using weighting schemes. \citet{liu2021towards} proposed impartial MTL that uses different strategies for task-shared parameters and task-specific parameters. \citet{pmlr-v162-navon22a} proposed to view the combination of gradients as a bargaining game, where different tasks negotiate with each other to reach an agreement on a joint direction of parameter update. They utilized the Nash Bargaining Solution~\citep{Nash1953} as an approach to address this problem and proved the effectiveness of their method across various tasks. Since some MTL methods are not always stable during training, \citet{Senushkin2023} proposed the Aligned MTL to improve stability. They used a condition number of a linear system of gradients as a stability criterion, and aligned the orthogonal components of the linear system of gradients to eliminate instability in training. 

The improved performance and stability of MTL methods have prompted its application to quality evaluation. \citet{shah-specia-2016-large} investigated MTL with Gaussian Processes for QE, based on datasets with multiple annotators and language pairs. They found multi-task models perform better than individual models in cross-lingual settings. \citet{zhang-van-genabith-2020-translation} used MTL to predict QE scores and rank different translations. \citet{rei-etal-2022-comet} employed MTL to jointly train QE models at sentence- and word-level. Most of these studies used non-parametric linear combinations of task losses, until~\citet{deoghare-etal-2023-multi} proposed to apply Nash MTL to combining sentence- and word-level QE, based on MicroTransQuest. However, their Nash MTL might not always be stable for various QE tasks. In this paper, we explore different MTL loss heuristics and propose a new architecture with a novel combined loss function for the quality estimation of emotion-loaded UGC. 

\section{Data} \label{data}

We used two datasets to evaluate our approach. The first one measures \textit{how well emotion is preserved} in machine translation and is presented in $\S$~\ref{hadqaet}. The second is a standard QE dataset from WMT 2020 to WMT 2022~\citep{freitag-etal-2021-experts, freitag-etal-2021-results, freitag-etal-2022-results}. It contains sentence- and word-level QE data annotated using MQM, as explained in $\S$~\ref{mqm_subset}. 

\subsection{A Human Annotated Dataset for Quality Assessment of Emotion Translation} \label{hadqaet}

We used our Human Annotated Dataset for Quality Assessment of Emotion Translation (HADQAET)\footnote{\url{https://github.com/surrey-nlp/HADQAET}} as the main resource~\citep{qian-etal-2023-evaluation}. Its source text originated from the dataset released by the~\textit{Evaluation of Weibo Emotion Classification Technology on the Ninth China National Conference on Social Media Processing} (SMP2020-EWECT). It originally has a size of 34,768 instances. Each instance is a tweet-like text segment\footnote{Like most NLP tasks, we treat tweet-like text segments as sentence-level data. However, in contrast to tweets, our instances are longer with an average of 40 Chinese characters.}, which was manually annotated in the original dataset with one of the six emotion labels, \textit{i.e.},~\textit{anger},~\textit{joy},~\textit{sadness},~\textit{surprise},~\textit{fear} and~\textit{neutral} ~\citep{Guo2021}. We kept 5,538 instances with labels other than \textit{neutral} and used Google Translate to translate them to English. We proposed an emotion-related MQM framework and recruited two professional translators to annotate errors and their corresponding severity in terms of emotion preservation. Words/characters in both source and target that cause errors were highlighted for error analysis. Details of our framework, error annotation (including inter-annotator agreement) and error analysis can be found in~\citet{qian-etal-2023-evaluation}. An example of the dataset is shown in Figure~\ref{fig.example2}. 

Since our original paper did not propose any scores for sentence-level QE, we followed~\citet{freitag-etal-2021-experts} to sum up all weighted errors based on their corresponding severity, using a set of weights\footnote{We validated these weights in~\citet{Qian2024}.} suggested by MQM~\citep{Lommel2014}, \textit{i.e.}, 1 for minor errors, 5 for major and 10 for critical. For word-level QE, we first tokenized the source with~\textit{jieba}~\citep{jieba2013}, and the target with NLTK~\citep{Bird2009} (tokenization tools for Chinese and English respectively). Then, we labeled the tokens highlighted by annotators as ``BAD'', and the rest ``OK''. This led to a sequence of labels for each instance, which indicate translation quality in emotion preservation at word level. 

The MQM-based QE scores related to emotion, word labels, together with the source texts and GT translations were used for quality estimation of emotion-loaded UGC. The emotion labels that were originally used for emotion classification were also incorporated to see if they are helpful for QE.

\subsection{MQM Subset with Synthetic Emotion} \label{mqm_subset}

\begin{figure}[h]
  \centering
  \includegraphics[width=0.45\textwidth]{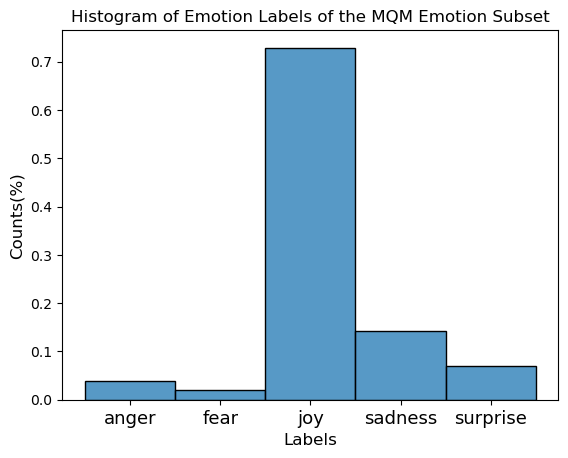}
  \caption{Distribution of the MQM emotion subset}
  \label{fig.subset-dis}
\end{figure}

To test whether our approach can be applied to standard QE data\footnote{Their QE scores are not related to emotion.}, we selected the overlapping of Chinese-English sentence- and word-level MQM datasets from the QE shared task of WMT 2020 to WMT 2022. The overlapped subset has MQM scores at sentence level and ``OK'' or ``BAD'' labels at word level. We fine-tuned the Chinese RoBERTa large model~\citep{cui-etal-2020-revisiting} on the SMP2020-EWECT dataset, resulting in an emotion classifier with a macro F1 score of 0.95. We predicted the emotion label of the source text of the selected data using the fine-tuned classifier, and filtered out all~\textit{neutral} instances. This led to an MQM subset with automatically generated emotion labels and a comparable size (3544) as HADQAET. We randomly sampled 100 instances and manually checked the predicted emotion labels with the help of a native speaker. The manual validation shows the emotion classifier is reliable as it achieves an F1 score 0.90, precision 0.91 and recall 0.92. The distribution of this subset is shown in Figure~\ref{fig.subset-dis}. 

\section{Methodology} \label{method}

This section describes the architecture and loss function of our MTL method. Additionally, it also presents the fine-tuning baselines including TransQuest and COMET for each individual task. 

\subsection{Multi-task Learning}

We propose a new architecture that is able to train sentence- and word-level QE systems with an emotion classifier using a combined loss function. 

\paragraph{Architecture}

\begin{figure}[h]
  \centering
  \includegraphics[width=0.45\textwidth]{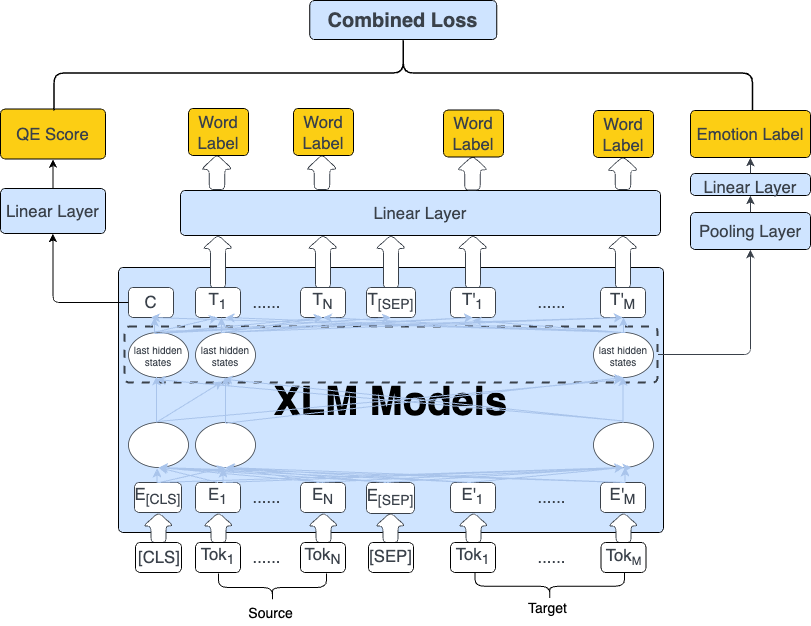}
  \caption{Architecture of our MTL Framework}
  \label{fig.archi}
\end{figure}

The architecture we propose is in Figure~\ref{fig.archi}. Following MonoTransQuest and COMETKIWI, we concatenate the source and target, including [CLS] and [SEP] as the starting and separating tokens. Then, we employed multilingual PTLMs like XLM-RoBERTa, XLM-V-base and InfoXLM~\citep{chi-etal-2021-infoxlm} to encode the input text. Different from~\citet{deoghare-etal-2023-multi}, who used embeddings of the last hidden layer, we utilized the output of the [CLS] token to predict sentence-level QE scores and the rest tokens for word label classification. On top of the encoder, we added a fully connected layer for both sentence- and word-level QE before the softmax function for prediction. 

To incorporate the emotion classification task,  we tried max and average pooling for the output of the last hidden layer of the encoder and added another fully connected layer on top. We used Xavier initialization~\citep{pmlr-v9-glorot10a} for the weights in all newly-added linear layers. We experimented different combination strategies for the losses of these tasks as explained below.

\paragraph{Combined Loss}

The loss function of our method is defined in Equation~\ref{eq:loss_mtl}, where $\sigma$ is a heuristic function to combine the three losses. $L_{sent}$ as shown in Equation~\ref{eq:loss_sent} is the Mean Squared Error loss for sentence-level QE, where $Y_{QE\_score}$ and $\hat{Y}_{QE\_score}$ are the true and predicted QE scores, respectively. Equation~\ref{eq:ce_loss} is the Cross Entropy loss for word and emotion classification, where $C$ is the set of classes. For $L_{word}$, $C$ is \{``OK'', ``BAD''\}. For $L_{emo}$, $C$ is the 5 emotion classes. 
$\mathbbm{1}\{y = i\}$ is an indicator function (1 if the true label $y$ is equal to the current class $i$, 0 otherwise), and $p_i$ is the predicted probability of the input being in class $i$.

\begin{equation} \label{eq:loss_mtl}
    L_{MTL} = \sigma(L_{sent}, L_{word}, L_{emo})
\end{equation} \\ [-9ex]

\begin{equation} \label{eq:loss_sent}
    L_{sent} = MSE(Y_{QE\_score}, \hat{Y}_{QE\_score})
\end{equation} \\ [-8.5ex]

\begin{equation} \label{eq:ce_loss}
    L_{word/emo} = -\sum_{i=1}^C \mathbbm{1}\{y = i\} \cdot \log(p_i)
\end{equation}

The objective of the heuristic $\sigma$ is to find a set of parameters $\theta$ that minimize the aggregate loss of all tasks. It is defined in Equation~\ref{eq:sigma}, where $L_{MTL}(\theta)$ is the combined loss, and $L_i(\theta)$ is the loss for an individual task $i$. 

\begin{equation} \label{eq:sigma}
    \theta^* = arg\ \underset{\theta}{min}\{L_{MTL}(\theta)=\Sigma^T_{i=1}L_i(\theta)\}
\end{equation}

Theoretically, $\theta$ can be fixed or a simple linear combination of each task loss. For instance, it can be 1 for each task loss, but the result is usually not ideal, as shown in our experiments. In order to balance the losses of different tasks and overcome optimization problems like conflicting or dominating gradients~\citep{{pmlr-v162-navon22a}}, we adopted different heuristics $\sigma$ to learn $\theta$, including the Nash and Aligned MTL losses which are explained in Appendix~\ref{sec:appendix}. Other existing MTL methods such as linear combination, dynamic weight averaging and impartial MTL were also integrated into our framework. To compare with our proposed Nash and Aligned MTL, the linear combination (1 for each task loss) and Nash MTL loss in~\citet{deoghare-etal-2023-multi} were selected as baseline MTL methods in our experiments. Results of other MTL methods are in Table~\ref{tab:other_loss}.

\subsection{Fine-tuning} \label{sec:finetuning}

We utilized MonoTransQuest, SiameseTransQuest and COMET for sentence-level QE, and MicroTransQuest for word-level QE. They rely on the XLM-RoBERTa models as the foundation model for fine-tuning. For emotion classification, we fine-tuned XLM-RoBERTa-large and XLM-V-base~\citep{Liang2023-hb} using both source and target texts. Experimental setup and training details can be seen in the following sections.

\subsection{Experimental Setup} \label{exp_setup}

We performed experiments under two settings (fine-tuning and MTL) on two datasets (HADQAET and the MQM emotion subset). Fine-tuning included sentence- and word-level QE and emotion classification. For MTL, we combined sentence-level QE with word-level QE, sentence-level QE with emotion classification, and sentence-, word-level QE and emotion classification. 

We used Spearman $\rho$ and Pearson's $r$ correlations to evaluate similarities between the predicted sentence-level QE scores and the true scores. For word and emotion classification, we used macro F1, precision and recall scores for evaluation.

\begin{table*}[h]
\centering
\resizebox{13cm}{!}{%
\begin{tabular}{ccccccc}
\toprule
\multicolumn{2}{c}{Methods} & \multicolumn{2}{c}{Sentence Level} & \multicolumn{3}{c}{Word Level} \\
Model & Loss & $\rho$ & $r$ & F & P & R \\ \hline
\multirow{4}{*}{XLM-RoBERTa-large} & Nash & 0.4024 & 0.3946 & 0.2664 & 0.2152 & \textbf{0.4055} \\
 & Aligned & 0.1214 & 0.1000 & 0.1835 & 0.1266 & 0.3333 \\
 & Linear & 0.1921 & 0.1779 & 0.1835 & 0.1266 & 0.3333 \\
 & Nash-D & 0.3642 & 0.3611 & 0.2465 & 0.1917 & 0.3885 \\ \cdashline{2-7}
\multirow{4}{*}{XLM-RoBERTa-base} & Nash & 0.2747 & 0.2589 & 0.2452 & 0.2126 & 0.3772 \\
 & Aligned & 0.2060 & 0.1629 & 0.1835 & 0.1266 & 0.3333 \\
 & Linear & 0.0354 & 0.0754 & 0.1835 & 0.1266 & 0.3333 \\
 &  Nash-D & 0.1278 & 0.1139 & 0.2565 & 0.2043 & 0.3844 \\ \cdashline{2-7}
\multirow{4}{*}{XLM-V-base} & Nash & \textbf{0.4673} & \textbf{0.4254} & \textbf{0.2805} & \textbf{0.2378} & 0.3953 \\
 & Aligned & 0.1391 & 0.1063 & 0.2538 & 0.2050 & 0.3333 \\ 
 & Linear & 0.2594 & 0.2052 & 0.2617 & 0.2154 & 0.3333 \\
 & Nash-D & 0.4290 & 0.3983 & 0.2495 & 0.1942 & 0.3923 \\ \hdashline
MicroTransQuest (FT) & / & / & / & 0.1951 & 0.6651 & 0.1143 \\
\bottomrule
\end{tabular}%
}
\caption{Spearman $\rho$, Pearson's $r$, Macro F1 (F), precision (P) and recall (R) scores of models combining sentence- and word-level QE using our MTL architecture~\textit{vs} other MTL methods including the linear loss and Nash loss from~\citet{deoghare-etal-2023-multi} (Nash-D) as well as the fine-tuning (FT) model using MicroTransQuest on HADQAET.}
\label{tab:mtl-sent-word}
\end{table*}

\subsection{Training Details} \label{training_details}

We divided the data into training, validation, and test sets in proportions of $80\%$, $10\%$, and $10\%$ respectively. We set the learning rate as $2e-5$ with the warmup rate as $0.1$, for all training setup. We chose the AdamW optimizer~\citep{Loshchilov2017-id} with a linear scheduler for all experiments. The sequence length was set as $200$ and the batch size was chosen as $8$. For fine-tuning, all models were trained for $2$ epochs except emotion classifiers; whereas for MTL, we trained our models for $8-12$ epochs based on the decrease of the combined loss and depending on different combinations of tasks. For the emotion classification task in MTL, we chose max pooling over average pooling after experimentation. We set the number of epochs as $10$ and used early stopping for fine-tuning emotion classifiers. All these hyperparameters were chosen based on experimentation and previous research. 

Fine-tuning multilingual PTLMs via TransQuest including MonoTransQuest, SiameseTransQuest and MicroTransQuest was carried out on an NVIDIA Quadro RTX 5000 GPU. Fine-tuning emotion classifiers including statistical models on HADQAET and the MQM emotion subset was performed on an NVIDIA T4 GPU. The rest of the model training including fine-tuning via COMET and different combinations of our MTL tasks were conducted on an NVIDIA A40 GPU.

\begin{figure}[h]
  \centering
  \includegraphics[width=0.45\textwidth]{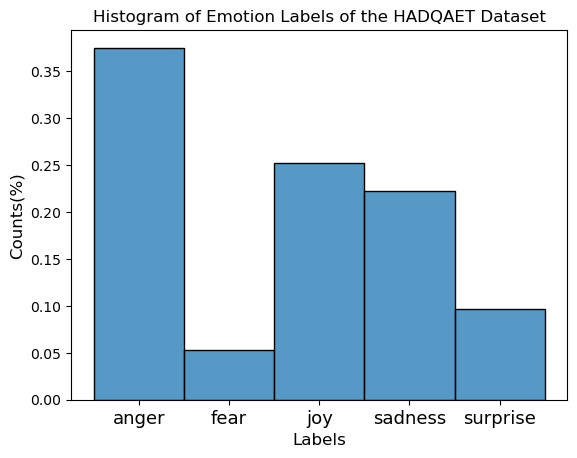}
  \caption{Distribution of the HADQAET dataset}
  \label{fig.hadqaet-dis}
\end{figure}

\begin{table}[h]
\centering
\resizebox{6cm}{!}{%
\begin{tabular}{lll}
\toprule
Methods & $\rho$ & $r$ \\ \hline
MonoTransQuest & \textbf{0.4355} & 0.3984 \\
SiameseTransQuest & 0.4151 & \textbf{0.4502} \\
COMET & 0.4083 & 0.3699 \\
\bottomrule
\end{tabular}%
}
\caption{Spearman $\rho$ and Pearson's $r$ correlation scores of models fine-tuned using TransQuest and COMET. }
\label{tab:ft}
\end{table}

\begin{table}[h]
\centering
\resizebox{7.7cm}{!}{%
\begin{tabular}{cccc}
\toprule
Methods & F & P & R \\ \hline
XLM-RoBERTa-large & 0.1000 & 0.0700 & 0.2000 \\
XLM-V-base & 0.1000 & 0.0700 & 0.2000 \\
RF on XLM-RoBERTa-large embeddings & \textbf{0.1456} & \textbf{0.1603} & \textbf{0.2072} \\
SVM on XLM-RoBERTa-large embeddings & 0.1169 & 0.0826 & 0.2000\\
\bottomrule
\end{tabular}%
}
\caption{Macro F1 (F), precision (P) and recall (R) scores of emotion classification models on HADQAET.}
\label{tab:classifier_hadqaet}
\end{table}

\begin{table*}[h]
\centering
\resizebox{13cm}{!}{%
\begin{tabular}{ccccccc}
\toprule
\multicolumn{2}{c}{Methods} & \multicolumn{2}{c}{Sentence Level} & \multicolumn{3}{c}{Emotion Classification} \\
Model & Loss & $\rho$ & $r$ & F & P & R \\ \hline
\multirow{3}{*}{XLM-RoBERTa-large} & Nash & \textit{-0.0357} & \textit{-0.0289} & 0.1073 & 0.0733 & 0.2000 \\
 & Aligned & 0.3786 & 0.3886 & 0.7985 & 0.7946 & 0.8257 \\
 & Linear & 0.2376 & 0.2715 & 0.8399 & 0.8263 & \textbf{0.8887} \\ \cdashline{2-7}
\multirow{3}{*}{XLM-RoBERTa-base} & Nash & 0.1448 & 0.1092 & \textbf{0.8549} & \textbf{0.8352} & 0.8879 \\
 & Aligned & 0.4229 & 0.4174 & 0.8198 & 0.8054 & 0.8510 \\
 & Linear & 0.3777 & 0.3521 & 0.7907 & 0.7756 & 0.8426 \\ \cdashline{2-7}
\multirow{3}{*}{XLM-V-base} & Nash & 0.0745 & 0.0105 & 0.1014  & 0.0679 & 0.2000 \\
 & Aligned & 0.4182 & 0.4278 & 0.8209 & 0.8040 & 0.8653 \\
 & Linear & -0.0621 & -0.0512 & 0.1014 & 0.0679 & 0.2000 \\ \cdashline{2-7}
 \multirow{1}{*}{FT baselines} & / & \textbf{0.4355} & \textbf{0.4502} & 0.1456 & 0.1603 & 0.2072 \\
 \bottomrule
\end{tabular}%
}
\caption{Spearman $\rho$, Pearson's $r$, Macro F1 (F), precision (P) and recall (R) scores of MTL models combining sentence-level QE and emotion classification using our MTL architecture~\textit{vs} linear loss on HADQAET. Our fine-tuning baselines (FT baselines) from Tables \ref{tab:ft} and \ref{tab:classifier_hadqaet} are listed here for reference.}
\label{tab:mtl_emo_sent_hadqaet}
\end{table*}

\section{Results and Discussion} \label{results}

The results obtained by different models are presented from $\S$~\ref{ft_hadqaet} to $\S$~\ref{results_subset}, while  $\S$~\ref{discussions} discusses the observations derived from our results.

\subsection{Fine-tuning on HADQAET} \label{ft_hadqaet}

This section shows the results of fine-tuning, the methods presented in $\S$~\ref{sec:finetuning} for sentence-level QE and emotion classification on HADQAET. The results at word-level QE are presented together with MTL in Table~\ref{tab:mtl-sent-word}.

Table~\ref{tab:ft} displays the results of sentence-level QE models on HADQAET. The highest correlation scores, $0.4355$ Spearman ($\rho$) and $0.4502$ Pearson ($r$), were achieved by fine-tuning using MonoTransQuest and SiameseTransQuest, respectively.

The emotion categories of HADQAET are imbalanced, and the dataset size is relatively small, as depicted in Figure~\ref{fig.hadqaet-dis}. As a result, the fine-tuned classifiers always predicted the same class. We tried different PTLMs and hyperparameters, but the performance was not better as seen in Table~\ref{tab:classifier_hadqaet}. For this reason, we applied statistical methods including Random Forest (RF)~\citep{Breiman2001} and Support Vector Machine (SVM)~\citep{Hearst1998} based on the embeddings from XLM-RoBERTa-large. Our baseline for emotion classification was established using RF, achieving the best F1 score of $0.1456$.

\subsection{MTL on HADQAET} \label{mtl_hadqaet}

This section shows results of different combinations of the three tasks on HADQAET.

\subsubsection{Sentence- and Word-level QE}

Table~\ref{tab:mtl-sent-word} shows results of MTL that combines sentence- and word-level QE. For sentence-level QE, it is observed that MTL using XLM-V-base and Nash loss achieved the highest $\rho$ of $0.4673$. This performance was superior to that of fine-tuning ($0.4355$). In the context of word-level QE, our best F1 score of $0.2805$ surpasses the performance of fine-tuning using MicroTransQuest, which achieved an F1 score of $0.1951$. This suggests that training sentence- and word-level QE systems together under the MTL framework can lead to improved performance in both tasks. Additionally, our MTL method is better than the linear loss and the Nash loss from~\citet{deoghare-etal-2023-multi} for both sentence- and word-level QE.

\subsubsection{Sentence-level QE and Emotion Classification} \label{sent_emo_hadqaet}

Table~\ref{tab:mtl_emo_sent_hadqaet} presents results for the combination of sentence-level QE and the emotion classification task. We can see that the use of MTL with Aligned loss effectively prevented the predictions from falling into the same category as shown in Table~\ref{tab:classifier_hadqaet}. Our top-performing model achieved an F1 score of $0.8549$, much higher than our baseline. Our Aligned loss usually performed better than the linear loss for both sentence-level QE and emotion classification. It appears that incorporating the sentence-level QE task has proven beneficial for training emotion classifiers. However, incorporating emotion classification does not seem to be very helpful for sentence-level QE, as Spearman scores are not higher than those of fine-tuned models. In addition, it has been observed that when combined with emotion classification, the Aligned loss demonstrates greater stability compared to the Nash loss. This method achieves a favorable equilibrium between sentence-level QE and emotion classification. 

\begin{table}[h]
\centering
\resizebox{7.5cm}{!}{%
\begin{tabular}{ccc}
\toprule
Heuristics & Sentence-level QE & Emotion Classification \\ \hline
Nash Loss & 0.5604 & 5.1199 \\
Aligned Loss & 0.6162 & 0.6377 \\
\bottomrule
\end{tabular}%
}
\caption{Average loss weights for sentence-level QE and emotion classification using Nash and Aligned losses}
\label{tab:loss_weights}
\end{table}

\begin{table*}[h]
\centering
\resizebox{14cm}{!}{%
\begin{tabular}{cccccccccc}
\toprule
\multicolumn{2}{c}{Methods} & \multicolumn{2}{c}{Sentence Level} & \multicolumn{3}{c}{Word Level} & \multicolumn{3}{c}{Emotion Classification} \\
Model & Loss & $\rho$ & $r$ & F & P & R & F & P & R \\ \hline
\multirow{3}{*}{XLM-RoBERTa-large} & Nash & 0.3787 & 0.3979 & 0.1735 & 0.2194 & 0.3805 & 0.8526 & \textbf{0.8419} & 0.8730 \\
 & Aligned & 0.1262 & 0.1035 & 0.1835 & 0.1266 & 0.3333 & 0.1014 & 0.0679 & 0.2000 \\
 & Linear & 0.4020 & 0.3573 & 0.1836 & 0.1267 & 0.3333 & 0.8159 & 0.8115 & 0.8625 \\ \cdashline{2-10}
\multirow{3}{*}{XLM-RoBERTa-base} & Nash & 0.2584 & 0.2342 & 0.2351 & 0.1740 & 0.3838 & \textbf{0.8528} & 0.8296 & 0.8903 \\
 & Aligned & 0.3786 & 0.3654 & 0.2013 & 0.1417 & 0.3472 & 0.8403 & 0.8185 & 0.8920 \\
 & Linear & 0.2895 & 0.2331 & 0.2131 & 0.1561 & 0.3426 & 0.7741 & 0.7658 & 0.8232 \\ \cdashline{2-10}
\multirow{3}{*}{XLM-V-base} & Nash & 0.4051 & 0.4082 & 0.2245 & 0.1631 & 0.3795 & 0.8513 & 0.8324 & \textbf{0.8938} \\
 & Aligned & 0.3389 & 0.3335 & 0.1914 & 0.1344 & 0.3337 & 0.8261 & 0.8220 & 0.8618 \\
 & Linear & 0.3610 & 0.3659 & \textbf{0.2461} & 0.2343 & \textbf{0.3992} & 0.7892 & 0.7740 & 0.8241 \\ \cdashline{2-10}
 \multirow{1}{*}{FT baselines} & / & \textbf{0.4355} & \textbf{0.4502} & 0.1951 & \textbf{0.6651} & 0.1143 & 0.1456 & 0.1603 & 0.2072 \\
 \bottomrule
\end{tabular}%
}
\caption{Spearman $\rho$, Pearson's $r$, Macro F1 (F), precision (P) and recall (R) scores of MTL models combining sentence- and word-level QE and emotion classification using our MTL architecture~\textit{vs} linear loss on HADQAET. Our fine-tuning baselines (FT baselines) from Tables \ref{tab:ft} and \ref{tab:classifier_hadqaet} are listed here for reference.}
\label{tab:mtl-all}
\end{table*}

\begin{table*}[h]
\centering
\resizebox{14cm}{!}{%
\begin{tabular}{ccccccccc}
\toprule
\multirow{2}{*}{Methods} & \multicolumn{2}{c}{Sentence Level} & \multicolumn{3}{c}{Word level} & \multicolumn{3}{c}{Emotion Classification} \\
 & $\rho$ & $r$ & F & P & R & F & P & R \\ \hline
MonoTransQuest & \textbf{0.3650} & \textbf{0.3836} & / & / & / & / & / & / \\
SiameseTransQuest & 0.2659 & 0.2622 & / & / & / & / & / & / \\
MicroTransQuest & / & / & \textbf{0.2141} & \textbf{0.4553} & \textbf{0.1399} & / & / & / \\
Random Forest & / & / & / & / & / & \textbf{0.1397} & \textbf{0.2061} & \textbf{0.2048} \\
SVM & / & / & / & / & / & 0.1202 & 0.0859 & 0.2000 \\
\bottomrule
\end{tabular}%
}
\caption{Spearman $\rho$, Pearson's $r$, Macro F1 (F), precision (P) and recall (R) scores for our baselines: fine-tuned models for sentence- and word-level QE and statistical models including Random Forest and Support Vector Machine (SVM) for emotion classification on the MQM emotion subset.}
\label{tab:ft_subset}
\end{table*}

Investigating further, we trained two models based on XLM-RoBERTa-base using the exact same hyperparameters, but two different loss heuristics\footnote{The linear loss was omitted as weights were fixed as 1.}, \textit{i.e.}, the Nash and Aligned losses, to combine sentence-level QE and emotion classification. We recorded the weights for the losses of the two tasks learned during training. The average loss weights (of all epochs) can be seen in Table~\ref{tab:loss_weights}. We can see that the Aligned loss seems to be better than Nash in balancing the two tasks as the two average weights are closer using the Aligned loss than Nash. This might be one of the reasons why it leads to more balanced results when the two tasks are combined.

\begin{table*}[h]
\centering
\resizebox{13cm}{!}{%
\begin{tabular}{ccccccc}
\toprule
\multicolumn{2}{c}{Methods} & \multicolumn{2}{c}{Sentence Level} & \multicolumn{3}{c}{Word Level} \\
Model & Loss & $\rho$ & $r$ & F & P & R \\ \hline
\multirow{4}{*}{XLM-RoBERTa-large} & Nash & 0.1212 & 0.2244 & 0.2437 & 0.1918 & 0.3996 \\
 & Aligned & 0.2840 & 0.2970 & 0.1682 & 0.1125 & 0.3333 \\
 & Linear & -0.1162 & -0.1249 & 0.1682 & 0.1125 & 0.3333 \\
 & Nash-D & 0.1427 & 0.1943 & 0.2447 & 0.1880 & \textbf{0.4043} \\ \cdashline{2-7}
\multirow{4}{*}{XLM-RoBERTa-base} & Nash & 0.1385 & 0.1157 & 0.2253 & 0.1781 & 0.3785 \\
 & Aligned & 0.2901 & 0.2928 & 0.1682 & 0.1125 & 0.3333 \\
 & Linear & 0.2250 & 0.2684 & 0.1682 & 0.1125 & 0.3333 \\
 & Nash-D & 0.2167 & 0.2304 & 0.2118 & 0.1549 & 0.3722 \\ \cdashline{2-7}
\multirow{4}{*}{XLM-V-base} & Nash & \textbf{0.4947} & \textbf{0.4448} & 0.2251 & 0.1603 & 0.3908 \\
 & Aligned & 0.3078 & 0.2204 & \textbf{0.2471} & 0.1963 & 0.3333 \\
 & Linear & 0.2635 & 0.2385 & 0.2465 & 0.1956 & 0.3333 \\
 & Nash-D & 0.1668 & 0.1619 & 0.2450 & 0.2057 & 0.3895 \\ \cdashline{2-7}
 \multirow{1}{*}{FT baselines} & / & 0.3650 & 0.3836 & 0.2141 & \textbf{0.4553} & 0.1399 \\
 \bottomrule
\end{tabular}%
}
\caption{Spearman $\rho$, Pearson's $r$, Macro F1 (F), precision (P) and recall (R) scores of models combining sentence- and word-level QE using our MTL architecture~\textit{vs} other MTL methods including the linear loss and Nash loss from~\citet{deoghare-etal-2023-multi} (Nash-D) on the MQM emotion subset. Our fine-tuning baselines (FT baselines) from Table \ref{tab:ft_subset} are listed here for reference.}
\label{tab:sent-word_subset}
\end{table*}

\subsubsection{Sentence-, Word-level QE and Emotion Classification}

\begin{table*}[h]
\centering
\resizebox{13cm}{!}{%
\begin{tabular}{ccccccc}
\toprule
\multicolumn{2}{c}{Methods} & \multicolumn{2}{c}{Sentence Level} & \multicolumn{3}{c}{Emotion Classification} \\
Model & Loss & $\rho$ & $r$ & F & P & R \\ \hline
\multirow{3}{*}{XLM-RoBERTa-large} & Nash & 0.3500 & 0.3737 & 0.0257 & 0.0265 & 0.0250 \\
 & Aligned & 0.1362 & 0.1699 & 0.1027 & 0.1014 & 0.1042 \\
 & Linear & 0.1593 & 0.0747 & 0.1742 & 0.1905 & 0.2689 \\ \cdashline{2-7}
\multirow{3}{*}{XLM-RoBERTa-base} & Nash & 0.1380 & 0.0125 & 0.1614 & 0.1595 & 0.2689 \\
 & Aligned & 0.1395 & 0.1684 & 0.1534 & 0.1239 & 0.2014 \\
 & Linear & 0.3305 & 0.3567 & 0.1273 & 0.1251 & 0.2106 \\ \cdashline{2-7}
\multirow{3}{*}{XLM-V-base} & Nash & 0.0631 & 0.0658 & 0.2185 & 0.1897 & 0.3409 \\
 & Aligned & \textit{-0.0894} & \textit{-0.0444} & \textbf{0.3004} & \textbf{0.2379} & \textbf{0.4862} \\
 & Linear & 0.0616 & 0.0058 & 0.1690 & 0.1723 & 0.2689 \\  \cdashline{2-7}
 \multirow{1}{*}{FT baselines} & / & \textbf{0.3650} & \textbf{0.3836} & 0.1397 & 0.2061 & 0.2048 \\
 \bottomrule
\end{tabular}%
}
\caption{Spearman $\rho$, Pearson's $r$, Macro F1 (F), precision (P) and recall (R) scores of models combining sentence-level QE and emotion classification tasks using our MTL architecture~\textit{vs} linear loss on the MQM emotion subset. Our fine-tuning baselines (FT baselines) from Table \ref{tab:ft_subset} are listed here for reference.}
\label{tab:sent-emo_subset}
\end{table*}

\begin{table*}[h]
\centering
\resizebox{14cm}{!}{%
\begin{tabular}{cccccccccc}
\toprule
\multicolumn{2}{c}{Methods} & \multicolumn{2}{c}{Sentence Level} & \multicolumn{3}{c}{Word Level} & \multicolumn{3}{c}{Emotion Classification} \\
Model & Loss & $\rho$ & $r$ & F & P & R & F & P & R \\ \hline
\multirow{3}{*}{XLM-RoBERTa-large} & Nash & 0.1198 & 0.1759 & 0.2284 & 0.1671 & \textbf{0.4116} & 0.1948 & 0.1623 & 0.2831 \\
 & Aligned & 0.1151 & 0.1613 & 0.1682 & 0.1125 & 0.3333 & 0.0553 & 0.0311 & 0.2500 \\
 & Linear & \textit{-0.1708} & \textit{-0.1581} & 0.1682 & 0.1125 & 0.3333 & 0.0553 & 0.0311 & 0.2500 \\ \cdashline{2-10}
\multirow{3}{*}{XLM-RoBERTa-base} & Nash & 0.2856 & \textit{-0.2112} & 0.2159 & 0.1523 & 0.4046 & 0.1392 & \textbf{0.3148} & 0.1935 \\
 & Aligned & 0.2878 & 0.2992 & \textbf{0.2497} & 0.2006 & 0.3306 & 0.1032 & 0.1074 & 0.1874 \\
 & Linear & 0.1794 & 0.1877 & 0.2151 & 0.1586 & 0.3447 & 0.1452 & 0.1661 & 0.2134 \\ \cdashline{2-10}
\multirow{3}{*}{XLM-V-base} & Nash & \textit{-0.0331} & 0.0392 & 0.1851 & 0.1383 & 0.3399 & 0.1520 & 0.1418 & 0.1755 \\
 & Aligned & \textbf{0.3779} & 0.2939 & 0.1736 & 0.1174 & 0.3333 & 0.1841 & 0.1592 & 0.2874 \\
 & Linear & 0.1130 & 0.1475 & 0.1743 & 0.1180 & 0.3333 & \textbf{0.2601} & 0.2120 & \textbf{0.4148} \\ \cdashline{2-10}
 \multirow{1}{*}{FT baselines} & / & 0.3650 & \textbf{0.3836} & 0.2141 & \textbf{0.4553} & 0.1399 & 0.1397 & 0.2061 & 0.2048 \\
 \bottomrule
\end{tabular}%
}
\caption{Spearman $\rho$, Pearson's $r$, Macro F1 (F), precision (P), recall (R) scores of models combining sentence- and word-level QE and emotion classification using our MTL architecture~\textit{vs} linear loss on the MQM emotion subset. Our fine-tuning baselines (FT baselines) from Table \ref{tab:ft_subset} are listed here for reference.}
\label{tab:mtl_all_subset}
\end{table*}

Table~\ref{tab:mtl-all} illustrates simultaneous training of the three tasks. Again, our MTL method achieved better results than the linear loss under most circumstances. Compared with fine-tuning, our MTL method notably enhanced the performance of emotion classification, but the result of sentence-level QE was compromised. This suggests that as more tasks are incorporated into the MTL framework, achieving consensus or agreement between tasks becomes more challenging. 

\subsection{Results on the MQM Emotion Subset} \label{results_subset}

This section presents results obtained on the MQM emotion subset, which is a selection of sentences from WMT QE shared tasks, with synthetic emotion labels as described in $\S$~\ref{mqm_subset}.

\subsubsection{Fine-tuning on MQM Emotion Subset} \label{ft_subset}

We applied the same methods as those of HADQAET, except that only statistical methods were used for emotion classification. Our baseline results are shown in Table~\ref{tab:ft_subset}. We achieved a $\rho$ of $0.3650$ for sentence-level QE, an F1 score of $0.2141$ for word-level QE and $0.1397$ for emotion classification. 

\subsubsection{MTL on MQM Emotion Subset} \label{mtl_subset}

Table~\ref{tab:sent-word_subset} presents the results of combining sentence- and word-level QE. Our best model, utilizing Nash loss, achieved a Spearman correlation of $0.4947$, notably surpassing the fine-tuning baseline and other MTL methods including the linear loss and Nash loss from~\citet{deoghare-etal-2023-multi}. The F1 score for word-level QE reached $0.2471$, demonstrating improvement over the fine-tuning baseline. These findings affirm the validity of our approach for effectively integrating sentence- and word-level QE in the context of overall quality evaluation. 

Table~\ref{tab:sent-emo_subset} shows results integrating sentence-level QE and emotion classification. In instances where sentence-level QE excelled ($\rho$ 0.35), we observed a trade-off with emotion classification performance, and vice versa. The use of the XLM-V base model with the Aligned loss improved the performance of emotion classification, resulting in the highest F1 score, 0.3004. 

Table~\ref{tab:mtl_all_subset} shows MTL results that combine all three tasks. Similar to results on HADQAET, there are trade-offs among tasks. Notably, on the MQM emotion subset, our best model achieved higher scores than fine-tuning and other MTL methods in both sentence- and word-level QE. This suggests that our approach contribute to the enhanced performance when training these tasks together.

\subsection{Discussion} \label{discussions}

The results obtained from various task combinations within our MTL framework indicate that training sentence- and word-level QE systems together improves their performance compared to training them separately. This improvement likely stems from the interconnected nature of the two QE tasks. However, adding emotion classification to the framework usually does not enhance sentence- or word-level QE. Conversely, combining sentence-level QE with emotion classification boosts the performance of emotion classification. This finding is consistent for both the HADQAET (an emotion-related QE dataset) and the MQM emotion subset (a standard QE dataset from WMT shared tasks). It suggests that the sentence-level QE task can aid in training emotion classifiers when training data is limited and the distribution is skewed. 

For word-level QE, our approach achieves higher recall scores than MicroTransQuest, possibly because our model predicts errors in both the source and target texts, whereas MicroTransQuest considers only errors in the target. 

Our results show that Nash and Aligned losses are generally better than the linear loss. Using the Nash loss is more likely to achieve state-of-the-art results for sentence-level QE, whereas the Aligned loss excels in balancing different tasks to produce a stable output. For this point, our observation still needs to be validated by further experiments on more task combinations and multilingual PTLMs. 

\section{Conclusion and Future Work} \label{conclusion}

To evaluate MT quality of emotion-loaded UGC at sentence- and word-level simultaneously, we employed an emotion-related dataset that includes emotion labels and human-annotated translation errors. We extended it with sentence-level QE scores and word labels. This led to a dataset suitable for sentence- and word-level QE, and emotion classification. We proposed a new architecture featuring a novel combined MTL loss function that integrates different loss heuristics. This approach unifies the training of multiple correlated tasks. We have made the code publicly available for similar task combinations such as empathy prediction and emotion classification. We compared our approach with existing fine-tuning and MTL methods and assessed its generalization on a standard QE dataset with synthetic emotion labels. We achieved new state-of-the-art results on both datasets. For future work, we aim to validate the effectiveness of our method on a larger multilingual QE dataset. We are also interested in investigating LLMs to evaluate machine translation of emotion-loaded UGC. 

\section{Limitations and Ethical Considerations} \label{limitations}

Although our MTL method is more effective, it is computationally expensive compared to fine-tuning for each task. Further, it takes longer to converge as parameters in the combined loss need to be learned over the training process. 

Incorporating emotion classification might lead to unstable performance for sentence-level QE under the Nash loss as explained in $\S$~\ref{sent_emo_hadqaet}. We will explore different task combinations and introduce a new hyperparameter to balance the tasks in our future work.

The experiments in the paper were conducted using publicly available datasets. New data were created based on those publicly available datasets using computer algorithms. No ethical approval was required as the use of all data in this paper follows the licenses in~\citet{qian-etal-2023-evaluation} and~\citet{freitag-etal-2021-experts, freitag-etal-2021-results, freitag-etal-2022-results}. 

\bibliography{anthology,custom}

\begin{thebibliography}{51}
\expandafter\ifx\csname natexlab\endcsname\relax\def\natexlab#1{#1}\fi

\bibitem[{Baxter(2000)}]{Baxter2000}
Jonathan Baxter. 2000.
\newblock {A Model of Inductive Bias Learning}.
\newblock \emph{J. Artif. Int. Res.}, 12(1):149–198.

\bibitem[{Bird et~al.(2009)Bird, Loper, and Klein}]{Bird2009}
Steven Bird, Edward Loper, and Ewan Klein. 2009.
\newblock \emph{Natural Language Processing with Python}.
\newblock O'Reilly Media Inc, Sebastopol, California.

\bibitem[{Blain et~al.(2023)Blain, Zerva, Ribeiro, Guerreiro, Kanojia, C.~de Souza, Silva, Vaz, Jingxuan, Azadi, Orasan, and Martins}]{blain-etal-2023-findings}
Frederic Blain, Chrysoula Zerva, Ricardo Ribeiro, Nuno~M. Guerreiro, Diptesh Kanojia, Jos{\'e}~G. C.~de Souza, Beatriz Silva, T{\^a}nia Vaz, Yan Jingxuan, Fatemeh Azadi, Constantin Orasan, and Andr{\'e} Martins. 2023.
\newblock \href {https://aclanthology.org/2023.wmt-1.52} {Findings of the {WMT} 2023 shared task on quality estimation}.
\newblock In \emph{Proceedings of the Eighth Conference on Machine Translation}, pages 629--653, Singapore. Association for Computational Linguistics.

\bibitem[{Breiman(2001)}]{Breiman2001}
Leo Breiman. 2001.
\newblock \href {https://doi.org/10.1023/A:1010933404324} {{Random Forests}}.
\newblock \emph{Machine Learning}, 45(1):5--32.

\bibitem[{Caruana(1997)}]{Caruana1997}
Rich Caruana. 1997.
\newblock \href {https://doi.org/10.1023/A:1007379606734} {{Multitask Learning}}.
\newblock \emph{Machine Learning}, 28:41--75.

\bibitem[{Chi et~al.(2021)Chi, Dong, Wei, Yang, Singhal, Wang, Song, Mao, Huang, and Zhou}]{chi-etal-2021-infoxlm}
Zewen Chi, Li~Dong, Furu Wei, Nan Yang, Saksham Singhal, Wenhui Wang, Xia Song, Xian-Ling Mao, Heyan Huang, and Ming Zhou. 2021.
\newblock \href {https://doi.org/10.18653/v1/2021.naacl-main.280} {{I}nfo{XLM}: An information-theoretic framework for cross-lingual language model pre-training}.
\newblock In \emph{Proceedings of the 2021 Conference of the North American Chapter of the Association for Computational Linguistics: Human Language Technologies}, pages 3576--3588, Online. Association for Computational Linguistics.

\bibitem[{Conneau et~al.(2020)Conneau, Khandelwal, Goyal, Chaudhary, Wenzek, Guzm{\'a}n, Grave, Ott, Zettlemoyer, and Stoyanov}]{conneau-etal-2020-unsupervised}
Alexis Conneau, Kartikay Khandelwal, Naman Goyal, Vishrav Chaudhary, Guillaume Wenzek, Francisco Guzm{\'a}n, Edouard Grave, Myle Ott, Luke Zettlemoyer, and Veselin Stoyanov. 2020.
\newblock \href {https://doi.org/10.18653/v1/2020.acl-main.747} {Unsupervised cross-lingual representation learning at scale}.
\newblock In \emph{Proceedings of the 58th Annual Meeting of the Association for Computational Linguistics}, pages 8440--8451, Online. Association for Computational Linguistics.

\bibitem[{Cui et~al.(2020)Cui, Che, Liu, Qin, Wang, and Hu}]{cui-etal-2020-revisiting}
Yiming Cui, Wanxiang Che, Ting Liu, Bing Qin, Shijin Wang, and Guoping Hu. 2020.
\newblock \href {https://doi.org/10.18653/v1/2020.findings-emnlp.58} {Revisiting pre-trained models for {C}hinese natural language processing}.
\newblock In \emph{Findings of the Association for Computational Linguistics: EMNLP 2020}, pages 657--668, Online. Association for Computational Linguistics.

\bibitem[{Deoghare et~al.(2023)Deoghare, Choudhary, Kanojia, Ranasinghe, Bhattacharyya, and Or{\u{a}}san}]{deoghare-etal-2023-multi}
Sourabh Deoghare, Paramveer Choudhary, Diptesh Kanojia, Tharindu Ranasinghe, Pushpak Bhattacharyya, and Constantin Or{\u{a}}san. 2023.
\newblock \href {https://doi.org/10.18653/v1/2023.findings-acl.585} {A multi-task learning framework for quality estimation}.
\newblock In \emph{Findings of the Association for Computational Linguistics: ACL 2023}, pages 9191--9205, Toronto, Canada. Association for Computational Linguistics.

\bibitem[{Dorr et~al.(2011)Dorr, Olive, McCary, and Christianson}]{Dorr2011}
Bonnie Dorr, Joseph Olive, John McCary, and Caitlin Christianson. 2011.
\newblock \href {https://doi.org/https://doi.org/10.1007/978-1-4419-7713-7_5} {{Machine Translation Evaluation and Optimization}}.
\newblock In J.~Olive, C.~Christianson, and J.~McCary, editors, \emph{Handbook of Natural Language Processing and Machine Translation}, pages 745--843. Springer.

\bibitem[{Fernandes et~al.(2023)Fernandes, Deutsch, Finkelstein, Riley, Martins, Neubig, Garg, Clark, Freitag, and Firat}]{fernandes-etal-2023-devil}
Patrick Fernandes, Daniel Deutsch, Mara Finkelstein, Parker Riley, Andr{\'e} Martins, Graham Neubig, Ankush Garg, Jonathan Clark, Markus Freitag, and Orhan Firat. 2023.
\newblock \href {https://doi.org/10.18653/v1/2023.wmt-1.100} {{The Devil Is in the Errors: Leveraging Large Language Models for Fine-grained Machine Translation Evaluation}}.
\newblock In \emph{Proceedings of the Eighth Conference on Machine Translation}, pages 1066--1083, Singapore. Association for Computational Linguistics.

\bibitem[{Freitag et~al.(2021{\natexlab{a}})Freitag, Foster, Grangier, Ratnakar, Tan, and Macherey}]{freitag-etal-2021-experts}
Markus Freitag, George Foster, David Grangier, Viresh Ratnakar, Qijun Tan, and Wolfgang Macherey. 2021{\natexlab{a}}.
\newblock \href {https://doi.org/10.1162/tacl_a_00437} {Experts, errors, and context: A large-scale study of human evaluation for machine translation}.
\newblock \emph{Transactions of the Association for Computational Linguistics}, 9:1460--1474.

\bibitem[{Freitag et~al.(2022)Freitag, Rei, Mathur, Lo, Stewart, Avramidis, Kocmi, Foster, Lavie, and Martins}]{freitag-etal-2022-results}
Markus Freitag, Ricardo Rei, Nitika Mathur, Chi-kiu Lo, Craig Stewart, Eleftherios Avramidis, Tom Kocmi, George Foster, Alon Lavie, and Andr{\'e} F.~T. Martins. 2022.
\newblock \href {https://aclanthology.org/2022.wmt-1.2} {Results of {WMT}22 metrics shared task: Stop using {BLEU} {--} neural metrics are better and more robust}.
\newblock In \emph{Proceedings of the Seventh Conference on Machine Translation (WMT)}, pages 46--68, Abu Dhabi, United Arab Emirates (Hybrid). Association for Computational Linguistics.

\bibitem[{Freitag et~al.(2021{\natexlab{b}})Freitag, Rei, Mathur, Lo, Stewart, Foster, Lavie, and Bojar}]{freitag-etal-2021-results}
Markus Freitag, Ricardo Rei, Nitika Mathur, Chi-kiu Lo, Craig Stewart, George Foster, Alon Lavie, and Ond{\v{r}}ej Bojar. 2021{\natexlab{b}}.
\newblock \href {https://aclanthology.org/2021.wmt-1.73} {Results of the {WMT}21 metrics shared task: Evaluating metrics with expert-based human evaluations on {TED} and news domain}.
\newblock In \emph{Proceedings of the Sixth Conference on Machine Translation}, pages 733--774, Online. Association for Computational Linguistics.

\bibitem[{Glorot and Bengio(2010)}]{pmlr-v9-glorot10a}
Xavier Glorot and Yoshua Bengio. 2010.
\newblock \href {https://proceedings.mlr.press/v9/glorot10a.html} {{Understanding the difficulty of training deep feedforward neural networks}}.
\newblock In \emph{Proceedings of the Thirteenth International Conference on Artificial Intelligence and Statistics}, volume~9 of \emph{Proceedings of Machine Learning Research}, pages 249--256, Chia Laguna Resort, Sardinia, Italy. PMLR.

\bibitem[{Graham et~al.(2013)Graham, Baldwin, Moffat, and Zobel}]{graham-etal-2013-continuous}
Yvette Graham, Timothy Baldwin, Alistair Moffat, and Justin Zobel. 2013.
\newblock \href {https://aclanthology.org/W13-2305} {Continuous measurement scales in human evaluation of machine translation}.
\newblock In \emph{Proceedings of the 7th Linguistic Annotation Workshop and Interoperability with Discourse}, pages 33--41, Sofia, Bulgaria. Association for Computational Linguistics.

\bibitem[{Guerreiro et~al.(2024)Guerreiro, Rei, Stigt, Coheur, Colombo, and Martins}]{Guerreiro2023}
Nuno~M. Guerreiro, Ricardo Rei, Daan~van Stigt, Luisa Coheur, Pierre Colombo, and André F.~T. Martins. 2024.
\newblock \href {https://doi.org/10.1162/tacl_a_00683} {{xcomet: Transparent Machine Translation Evaluation through Fine-grained Error Detection}}.
\newblock \emph{Transactions of the Association for Computational Linguistics}, 12:979--995.

\bibitem[{Guo et~al.(2021)Guo, Lai, Xiang, Yu, and Huang}]{Guo2021}
Xianwei Guo, Hua Lai, Yan Xiang, Zhengtao Yu, and Yuxin Huang. 2021.
\newblock \href {https://aclanthology.org/2021.ccl-1.82} {{Emotion Classification of COVID-19 Chinese Microblogs based on the Emotion Category Description}}.
\newblock In \emph{Proceedings of the 20th China National Conference on Computational Linguistics}, pages 916--927. Chinese Information Processing Society of China.

\bibitem[{Hassan et~al.(2018)Hassan, Aue, Chen, Chowdhary, Clark, Federmann, Huang, Junczys-Dowmunt, Lewis, Li, Liu, Liu, Luo, Menezes, Qin, Seide, Tan, Tian, Wu, Wu, Xia, Zhang, Zhang, and Zhou}]{Hassan2018}
Hany Hassan, Anthony Aue, Chang Chen, Vishal Chowdhary, Jonathan Clark, Christian Federmann, Xuedong Huang, Marcin Junczys-Dowmunt, William Lewis, Mu~Li, Shujie Liu, Tie-Yan Liu, Renqian Luo, Arul Menezes, Tao Qin, Frank Seide, Xu~Tan, Fei Tian, Lijun Wu, Shuangzhi Wu, Yingce Xia, Dongdong Zhang, Zhirui Zhang, and Ming Zhou. 2018.
\newblock \href {http://arxiv.org/abs/arXiv:1803.05567} {{Achieving Human Parity on Automatic Chinese to English News Translation}}.
\newblock \emph{arXive preprint}.

\bibitem[{Hearst et~al.(1998)Hearst, Dumais, Osuna, Platt, and Scholkopf}]{Hearst1998}
M.A. Hearst, S.T. Dumais, E.~Osuna, J.~Platt, and B.~Scholkopf. 1998.
\newblock \href {https://doi.org/10.1109/5254.708428} {{Support vector machines}}.
\newblock \emph{IEEE Intelligent Systems and their Applications}, 13(4):18--28.

\bibitem[{Kepler et~al.(2019)Kepler, Tr{\'e}nous, Treviso, Vera, and Martins}]{kepler-etal-2019-openkiwi}
Fabio Kepler, Jonay Tr{\'e}nous, Marcos Treviso, Miguel Vera, and Andr{\'e} F.~T. Martins. 2019.
\newblock \href {https://doi.org/10.18653/v1/P19-3020} {{O}pen{K}iwi: An open source framework for quality estimation}.
\newblock In \emph{Proceedings of the 57th Annual Meeting of the Association for Computational Linguistics: System Demonstrations}, pages 117--122, Florence, Italy. Association for Computational Linguistics.

\bibitem[{Kocmi and Federmann(2023{\natexlab{a}})}]{kocmi-federmann-2023-gemba}
Tom Kocmi and Christian Federmann. 2023{\natexlab{a}}.
\newblock \href {https://doi.org/10.18653/v1/2023.wmt-1.64} {{GEMBA}-{MQM}: Detecting translation quality error spans with {GPT}-4}.
\newblock In \emph{Proceedings of the Eighth Conference on Machine Translation}, pages 768--775, Singapore. Association for Computational Linguistics.

\bibitem[{Kocmi and Federmann(2023{\natexlab{b}})}]{kocmi-federmann-2023-large}
Tom Kocmi and Christian Federmann. 2023{\natexlab{b}}.
\newblock \href {https://aclanthology.org/2023.eamt-1.19} {Large language models are state-of-the-art evaluators of translation quality}.
\newblock In \emph{Proceedings of the 24th Annual Conference of the European Association for Machine Translation}, pages 193--203, Tampere, Finland. European Association for Machine Translation.

\bibitem[{Lai et~al.(2020)Lai, Dai, and Yang}]{Lai2020}
Guokun Lai, Zihang Dai, and Yiming Yang. 2020.
\newblock \href {http://arxiv.org/abs/2009.08595} {{Unsupervised Parallel Corpus Mining on Web Data}}.
\newblock \emph{arXiv preprint}.

\bibitem[{Liang et~al.(2023)Liang, Gonen, Mao, Hou, Goyal, Ghazvininejad, Zettlemoyer, and Khabsa}]{Liang2023-hb}
Davis Liang, Hila Gonen, Yuning Mao, Rui Hou, Naman Goyal, Marjan Ghazvininejad, Luke Zettlemoyer, and Madian Khabsa. 2023.
\newblock \href {https://doi.org/10.18653/v1/2023.emnlp-main.813} {{XLM-V: Overcoming the Vocabulary Bottleneck in Multilingual Masked Language Models}}.
\newblock In \emph{Proceedings of the 2023 Conference on Empirical Methods in Natural Language Processing}, pages 13142--13152, Singapore. Association for Computational Linguistics.

\bibitem[{Liu et~al.(2021)Liu, Li, Kuang, Xue, Chen, Yang, Liao, and Zhang}]{liu2021towards}
Liyang Liu, Yi~Li, Zhanghui Kuang, Jing-Hao Xue, Yimin Chen, Wenming Yang, Qingmin Liao, and Wayne Zhang. 2021.
\newblock \href {https://openreview.net/forum?id=IMPnRXEWpvr} {{Towards Impartial Multi-task Learning}}.
\newblock In \emph{International Conference on Learning Representations}.

\bibitem[{Liu et~al.(2019)Liu, Johns, and Davison}]{Liu2019-dwa}
S.~Liu, E.~Johns, and A.~J. Davison. 2019.
\newblock \href {https://doi.org/10.1109/CVPR.2019.00197} {{End-To-End Multi-Task Learning With Attention}}.
\newblock In \emph{2019 IEEE/CVF Conference on Computer Vision and Pattern Recognition (CVPR)}, pages 1871--1880, Los Alamitos, CA, USA. IEEE Computer Society.

\bibitem[{Lommel et~al.(2014)Lommel, Burchardt, and Uszkoreit}]{Lommel2014}
Arle~Richard Lommel, Aljoscha Burchardt, and Hans Uszkoreit. 2014.
\newblock \href {https://doi.org/10.5565/rev/tradumatica.77} {{Multidimensional Quality Metrics: A Flexible System for Assessing Translation Quality}}.
\newblock \emph{Tradumàtica: tecnologies de la traducció}, 0:455--463.

\bibitem[{Loshchilov and Hutter(2019)}]{Loshchilov2017-id}
Ilya Loshchilov and Frank Hutter. 2019.
\newblock \href {https://openreview.net/forum?id=Bkg6RiCqY7} {{Decoupled Weight Decay Regularization}}.
\newblock In \emph{International Conference on Learning Representations}.

\bibitem[{Nash(1953)}]{Nash1953}
John Nash. 1953.
\newblock \href {http://www.jstor.org/stable/1906951} {{Two-Person Cooperative Games}}.
\newblock \emph{Econometrica}, 21(1):128--140.

\bibitem[{Navon et~al.(2022)Navon, Shamsian, Achituve, Maron, Kawaguchi, Chechik, and Fetaya}]{pmlr-v162-navon22a}
Aviv Navon, Aviv Shamsian, Idan Achituve, Haggai Maron, Kenji Kawaguchi, Gal Chechik, and Ethan Fetaya. 2022.
\newblock \href {https://proceedings.mlr.press/v162/navon22a.html} {{Multi-Task Learning as a Bargaining Game}}.
\newblock In \emph{Proceedings of the 39th International Conference on Machine Learning}, volume 162 of \emph{Proceedings of Machine Learning Research}, pages 16428--16446. PMLR.

\bibitem[{Peters et~al.(2018)Peters, Neumann, Iyyer, Gardner, Clark, Lee, and Zettlemoyer}]{peters-etal-2018-deep}
Matthew~E. Peters, Mark Neumann, Mohit Iyyer, Matt Gardner, Christopher Clark, Kenton Lee, and Luke Zettlemoyer. 2018.
\newblock \href {https://doi.org/10.18653/v1/N18-1202} {Deep contextualized word representations}.
\newblock In \emph{Proceedings of the 2018 Conference of the North {A}merican Chapter of the Association for Computational Linguistics: Human Language Technologies, Volume 1 (Long Papers)}, pages 2227--2237, New Orleans, Louisiana. Association for Computational Linguistics.

\bibitem[{Qian et~al.(2023)Qian, Orasan, Carmo, Li, and Kanojia}]{qian-etal-2023-evaluation}
Shenbin Qian, Constantin Orasan, Felix~Do Carmo, Qiuliang Li, and Diptesh Kanojia. 2023.
\newblock \href {https://aclanthology.org/2023.eamt-1.13} {Evaluation of {C}hinese-{E}nglish machine translation of emotion-loaded microblog texts: A human annotated dataset for the quality assessment of emotion translation}.
\newblock In \emph{Proceedings of the 24th Annual Conference of the European Association for Machine Translation}, pages 125--135, Tampere, Finland. European Association for Machine Translation.

\bibitem[{Qian et~al.(2024)Qian, Orasan, Kanojia, and Do~Carmo}]{Qian2024}
Shenbin Qian, Constantin Orasan, Diptesh Kanojia, and F{\'e}lix Do~Carmo. 2024.
\newblock {Are Large Language Models State-of-the-art Quality Estimators for Machine Translation of User-generated Content?}
\newblock In \emph{Proceedings of the 11th Workshop on Asian Translation}, Miami, United States of America. Association for Computational Linguistics.

\bibitem[{Ranasinghe et~al.(2020)Ranasinghe, Orasan, and Mitkov}]{ranasinghe-etal-2020-transquest}
Tharindu Ranasinghe, Constantin Orasan, and Ruslan Mitkov. 2020.
\newblock \href {https://doi.org/10.18653/v1/2020.coling-main.445} {{T}rans{Q}uest: Translation quality estimation with cross-lingual transformers}.
\newblock In \emph{Proceedings of the 28th International Conference on Computational Linguistics}, pages 5070--5081, Barcelona, Spain (Online). International Committee on Computational Linguistics.

\bibitem[{Ranasinghe et~al.(2021)Ranasinghe, Orasan, and Mitkov}]{ranasinghe-etal-2021-exploratory}
Tharindu Ranasinghe, Constantin Orasan, and Ruslan Mitkov. 2021.
\newblock \href {https://doi.org/10.18653/v1/2021.acl-short.55} {An exploratory analysis of multilingual word-level quality estimation with cross-lingual transformers}.
\newblock In \emph{Proceedings of the 59th Annual Meeting of the Association for Computational Linguistics and the 11th International Joint Conference on Natural Language Processing (Volume 2: Short Papers)}, pages 434--440, Online. Association for Computational Linguistics.

\bibitem[{Rei et~al.(2022{\natexlab{a}})Rei, C.~de Souza, Alves, Zerva, Farinha, Glushkova, Lavie, Coheur, and Martins}]{rei-etal-2022-comet}
Ricardo Rei, Jos{\'e}~G. C.~de Souza, Duarte Alves, Chrysoula Zerva, Ana~C Farinha, Taisiya Glushkova, Alon Lavie, Luisa Coheur, and Andr{\'e} F.~T. Martins. 2022{\natexlab{a}}.
\newblock \href {https://aclanthology.org/2022.wmt-1.52} {{COMET}-22: Unbabel-{IST} 2022 submission for the metrics shared task}.
\newblock In \emph{Proceedings of the Seventh Conference on Machine Translation (WMT)}, pages 578--585, Abu Dhabi, United Arab Emirates (Hybrid). Association for Computational Linguistics.

\bibitem[{Rei et~al.(2020)Rei, Stewart, Farinha, and Lavie}]{rei-etal-2020-comet}
Ricardo Rei, Craig Stewart, Ana~C Farinha, and Alon Lavie. 2020.
\newblock \href {https://doi.org/10.18653/v1/2020.emnlp-main.213} {{COMET}: A neural framework for {MT} evaluation}.
\newblock In \emph{Proceedings of the 2020 Conference on Empirical Methods in Natural Language Processing (EMNLP)}, pages 2685--2702, Online. Association for Computational Linguistics.

\bibitem[{Rei et~al.(2022{\natexlab{b}})Rei, Treviso, Guerreiro, Zerva, Farinha, Maroti, C.~de Souza, Glushkova, Alves, Coheur, Lavie, and Martins}]{rei-etal-2022-cometkiwi}
Ricardo Rei, Marcos Treviso, Nuno~M. Guerreiro, Chrysoula Zerva, Ana~C Farinha, Christine Maroti, Jos{\'e}~G. C.~de Souza, Taisiya Glushkova, Duarte Alves, Luisa Coheur, Alon Lavie, and Andr{\'e} F.~T. Martins. 2022{\natexlab{b}}.
\newblock \href {https://aclanthology.org/2022.wmt-1.60} {{C}omet{K}iwi: {IST}-unbabel 2022 submission for the quality estimation shared task}.
\newblock In \emph{Proceedings of the Seventh Conference on Machine Translation (WMT)}, pages 634--645, Abu Dhabi, United Arab Emirates (Hybrid). Association for Computational Linguistics.

\bibitem[{Saadany et~al.(2023)Saadany, Orasan, Quintana, Carmo, and Zilio}]{saadany-etal-2023-analysing}
Hadeel Saadany, Constantin Orasan, Rocio~Caro Quintana, Felix~Do Carmo, and Leonardo Zilio. 2023.
\newblock \href {https://aclanthology.org/2023.eamt-1.27} {Analysing mistranslation of emotions in multilingual tweets by online {MT} tools}.
\newblock In \emph{Proceedings of the 24th Annual Conference of the European Association for Machine Translation}, pages 275--284, Tampere, Finland. European Association for Machine Translation.

\bibitem[{Senushkin et~al.(2023)Senushkin, Patakin, Kuznetsov, and Konushin}]{Senushkin2023}
D.~Senushkin, N.~Patakin, A.~Kuznetsov, and A.~Konushin. 2023.
\newblock \href {https://doi.org/10.1109/CVPR52729.2023.01923} {{Independent Component Alignment for Multi-Task Learning}}.
\newblock In \emph{2023 IEEE/CVF Conference on Computer Vision and Pattern Recognition (CVPR)}, pages 20083--20093, Los Alamitos, CA, USA. IEEE Computer Society.

\bibitem[{Shah and Specia(2016)}]{shah-specia-2016-large}
Kashif Shah and Lucia Specia. 2016.
\newblock \href {https://doi.org/10.18653/v1/N16-1069} {Large-scale multitask learning for machine translation quality estimation}.
\newblock In \emph{Proceedings of the 2016 Conference of the North {A}merican Chapter of the Association for Computational Linguistics: Human Language Technologies}, pages 558--567, San Diego, California. Association for Computational Linguistics.

\bibitem[{Snover et~al.(2006)Snover, Dorr, Schwartz, Micciulla, and Makhoul}]{snover-etal-2006-study}
Matthew Snover, Bonnie Dorr, Rich Schwartz, Linnea Micciulla, and John Makhoul. 2006.
\newblock \href {https://aclanthology.org/2006.amta-papers.25} {A study of translation edit rate with targeted human annotation}.
\newblock In \emph{Proceedings of the 7th Conference of the Association for Machine Translation in the Americas: Technical Papers}, pages 223--231, Cambridge, Massachusetts, USA. Association for Machine Translation in the Americas.

\bibitem[{Specia et~al.(2020)Specia, Blain, Fomicheva, Fonseca, Chaudhary, Guzm{\'a}n, and Martins}]{specia-etal-2020-findings-wmt}
Lucia Specia, Fr{\'e}d{\'e}ric Blain, Marina Fomicheva, Erick Fonseca, Vishrav Chaudhary, Francisco Guzm{\'a}n, and Andr{\'e} F.~T. Martins. 2020.
\newblock \href {https://aclanthology.org/2020.wmt-1.79} {Findings of the {WMT} 2020 shared task on quality estimation}.
\newblock In \emph{Proceedings of the Fifth Conference on Machine Translation}, pages 743--764, Online. Association for Computational Linguistics.

\bibitem[{Specia et~al.(2021)Specia, Blain, Fomicheva, Zerva, Li, Chaudhary, and Martins}]{specia-etal-2021-findings}
Lucia Specia, Fr{\'e}d{\'e}ric Blain, Marina Fomicheva, Chrysoula Zerva, Zhenhao Li, Vishrav Chaudhary, and Andr{\'e} F.~T. Martins. 2021.
\newblock \href {https://aclanthology.org/2021.wmt-1.71} {Findings of the {WMT} 2021 shared task on quality estimation}.
\newblock In \emph{Proceedings of the Sixth Conference on Machine Translation}, pages 684--725, Online. Association for Computational Linguistics.

\bibitem[{Standley et~al.(2020)Standley, Zamir, Chen, Guibas, Malik, and Savarese}]{Standley2020}
Trevor Standley, Amir Zamir, Dawn Chen, Leonidas Guibas, Jitendra Malik, and Silvio Savarese. 2020.
\newblock {Which Tasks Should Be Learned Together in Multi-Task Learning?}
\newblock In \emph{Proceedings of the 37th International Conference on Machine Learning}, ICML'20. JMLR.org.

\bibitem[{Stewart et~al.(2020)Stewart, Rei, Farinha, and Lavie}]{stewart-etal-2020-comet}
Craig Stewart, Ricardo Rei, Catarina Farinha, and Alon Lavie. 2020.
\newblock \href {https://aclanthology.org/2020.amta-user.4} {{COMET} - deploying a new state-of-the-art {MT} evaluation metric in production}.
\newblock In \emph{Proceedings of the 14th Conference of the Association for Machine Translation in the Americas (Volume 2: User Track)}, pages 78--109, Virtual. Association for Machine Translation in the Americas.

\bibitem[{Sun(2013)}]{jieba2013}
Andy Sun. 2013.
\newblock Jieba.
\newblock \url{https://github.com/fxsjy/jieba}.

\bibitem[{Wang et~al.(2021)Wang, Tu, Tan, Wang, Sun, and Liu}]{Wang2021-lh}
Shuo Wang, Zhaopeng Tu, Zhixing Tan, Wenxuan Wang, Maosong Sun, and Yang Liu. 2021.
\newblock \href {http://arxiv.org/abs/2106.13627} {{Language Models are Good Translators}}.
\newblock \emph{arXiv preprint}.

\bibitem[{Zerva et~al.(2022)Zerva, Blain, Rei, Lertvittayakumjorn, C.~de Souza, Eger, Kanojia, Alves, Or{\u{a}}san, Fomicheva, Martins, and Specia}]{zerva-etal-2022-findings}
Chrysoula Zerva, Fr{\'e}d{\'e}ric Blain, Ricardo Rei, Piyawat Lertvittayakumjorn, Jos{\'e}~G. C.~de Souza, Steffen Eger, Diptesh Kanojia, Duarte Alves, Constantin Or{\u{a}}san, Marina Fomicheva, Andr{\'e} F.~T. Martins, and Lucia Specia. 2022.
\newblock \href {https://aclanthology.org/2022.wmt-1.3} {Findings of the {WMT} 2022 shared task on quality estimation}.
\newblock In \emph{Proceedings of the Seventh Conference on Machine Translation (WMT)}, pages 69--99, Abu Dhabi, United Arab Emirates (Hybrid). Association for Computational Linguistics.

\bibitem[{Zhang and van Genabith(2020)}]{zhang-van-genabith-2020-translation}
Jingyi Zhang and Josef van Genabith. 2020.
\newblock \href {https://doi.org/10.18653/v1/2020.emnlp-main.205} {Translation quality estimation by jointly learning to score and rank}.
\newblock In \emph{Proceedings of the 2020 Conference on Empirical Methods in Natural Language Processing (EMNLP)}, pages 2592--2598, Online. Association for Computational Linguistics.

\end{thebibliography}

\clearpage
\appendix

\section{Appendix} \label{sec:appendix}
\counterwithin{figure}{section}
\setcounter{figure}{0} 
\counterwithin{table}{section}
\setcounter{table}{0} 

\subsection{Additional Figures and Tables}

\begin{figure*}[h]
  \centering
  \includegraphics[width=0.99\textwidth]{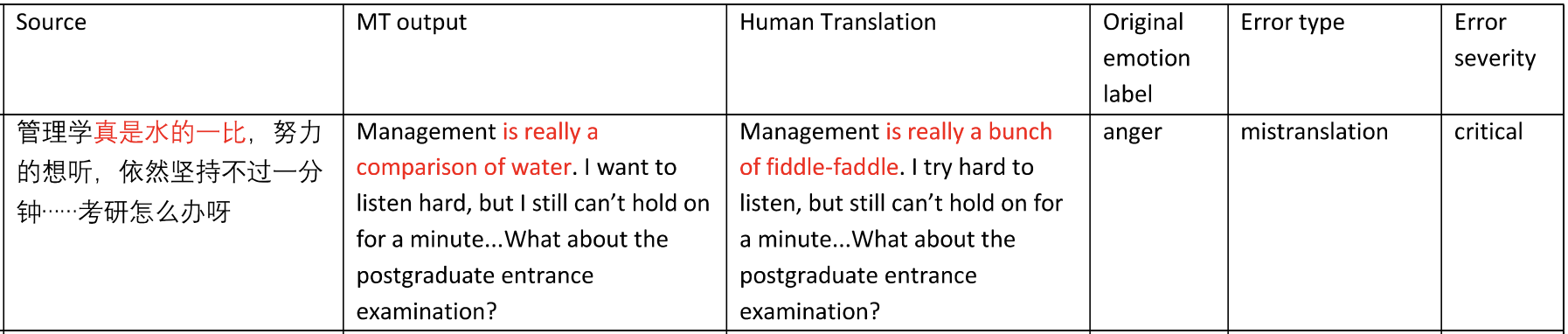}
  \caption{An Example from HADQAET~\citep{qian-etal-2023-evaluation}}
  \label{fig.example2}
\end{figure*}

Figure \ref{fig.example2} shows an example of the HADQAET dataset from \citet{qian-etal-2023-evaluation}. Table \ref{tab:other_loss} displays results of other loss heuristics in our framework.

\begin{table*}[h]
\centering
\resizebox{16cm}{!}{
\begin{tabular}{ccccccc}
\toprule
\multicolumn{2}{c}{Methods} & \multicolumn{2}{c}{Sentence Level} & \multicolumn{3}{c}{Word Level} \\
Model & Loss & $\rho$ & $r$ & F & P & R \\ \hline
\multirow{2}{*}{XLM-RoBERTa-large} & DWA & -0.0740 & -0.1031 & 0.1835 & 0.1266 & 0.3333 \\
 & IMTL & 0.1488 & 0.1057 & 0.2440 & 0.2096 & 0.3767 \\ \cdashline{2-7}
 \multirow{2}{*}{XLM-RoBERTa-base} & DWA & 0.0533 & 0.0726 & 0.0183 & 0.0094 & 0.3333 \\
 & IMTL & 0.1495 & 0.1561 & 0.2322 & 0.1929 & 0.3668 \\ \cdashline{2-7}
  \multirow{2}{*}{XLM-V-base} & DWA & -0.2551 & -0.2302 & 0.1870 & 0.1300 & 0.3333 \\
 & IMTL & 0.3182 & 0.2714 & 0.2757 & 0.2320 & 0.3843 \\ 
  \hdashline
\multirow{6}{*}{InfoXLM} & Nash & 0.1678 & 0.2647 & 0.2454 & 0.2181 & 0.3763 \\
 & Aligned & 0.0363 & 0.0281 & 0.1835 &0.1266 & 0.3333 \\
 & DWA & -0.0237 & -0.0355 & 0.1835 & 0.1266 & 0.3333 \\
 & IMTL & -0.2731 & -0.2200 & 0.1879 & 0.1941 & 0.3353 \\
 & Linear & 0.0042 & 0.0013 & 0.1835 & 0.1266 & 0.3333 \\
 & Nash-D & 0.1846 & 0.2125 & 0.2618 & 0.2377 & 0.3902 \\
 \bottomrule
\end{tabular}%
}
\caption{Spearman $\rho$, Pearson's $r$, Macro F1 (F), precision (P) and recall (R) scores of models fine-tuned based on XLM-RoBERTa, XLM-V-base and InfoXLM models in combination of sentence- and word-level QE using Dynamic Weight Averaging (DWA) and impartial MTL (IMTL) on HADQAET. Results obtained using the linear combination and Nash MTL in~\citet{deoghare-etal-2023-multi}, \textit{i.e.}, Nash-D, for InfoXLM are also displayed here.}
\label{tab:other_loss}
\end{table*}

\subsection{Nash MTL}

Nash MTL intends to find an update vector $\Delta\theta$ for the gradients $g_i$ of the task $i$ in the ball of radius $\epsilon$ centered around zero, $B_\epsilon$, as shown in Equation~\ref{eq:nash}.

\begin{equation} \label{eq:nash}
    arg\ max_{\Delta\theta\in B_{\epsilon}} \Sigma_i log(\Delta\theta^\intercal g_i)
\end{equation}

The solution to Equation~\ref{eq:nash} is (up to scaling) $\Sigma_i \alpha_i g_i$ where $\alpha \in \mathbb{R}_+^K$ is the solution to $G^\intercal G \alpha = 1 / \alpha$ where $1 / \alpha$ is the element-wise reciprocal. Detailed proof can be seen in~\citet{pmlr-v162-navon22a}. The Nash MTL algorithm is shown below:

\begin{table}[H]
\centering
\resizebox{7.7cm}{!}{%
\begin{tabular}{l}
\toprule
Algorithm 1 Nash-MTL \\ \hline
\textbf{Input}:  $\theta^{(0)}$ – initial parameter vector, $\{l_i\}^K_{i=1}$ – \\ differentiable loss functions $\eta$ – learning rate\\
\textbf{for} $t = 1,...,T$ \textbf{do} \\
\ \ Compute task gradients $g_i^{(t)}=\nabla_{\theta ^{(t-1)}}l_i$\\
\ \ Set $G^{(t)}$ the matrix with columns $g_i^{(t)}$ \\
\ \ Solve for $\alpha: (G^t)^\intercal G \alpha = 1 / \alpha$ to obtain $\alpha^{(t)}$ \\
\ \ Update the parameters  $\theta^{(t)} = \theta^{(t)} - \eta G^{(t)} \alpha^{(t)}$ \\
\textbf{end for} \\
\textbf{Return} $\theta^{(T)}$ \\
\bottomrule
\end{tabular}%
}
\label{tab:nash_algo}
\end{table}

\subsection{Aligned MTL}

Through theoretical analysis, \citet{Senushkin2023} found a strong relation between the condition number and conflicting and dominating gradients issues, and they proposed Aligned MTL to align principal components of a gradient matrix to make the training process more stable. 

The objective of Aligned MTL as defined in Equation~\ref{eq:aligned-obj}, is to minimize the difference between the original gradient matrix $G$ and the aligned gradient matrix $\hat{G}$. The difference is measured using the Frobenius $F$ norm. The constraint in Equation~\ref{eq:aligned-obj} ensures that $\hat{G}$ is orthogonal, meaning that its transpose multiplied by itself is equal to the identity matrix. This constraint helps to ensure stability in the linear system of gradients. 

\begin{equation} \label{eq:aligned-obj}
    \underset{\hat{G}}{min} \lVert G- \hat{G} \rVert^2_F \ \ s.t.\ \ \hat{G}^\intercal \hat{G} = I
\end{equation}

\begin{equation} \label{eq:aligned-algo}
    \hat{G} = \sigma UV^\intercal = \sigma GV \Sigma^{-1}V^\intercal
\end{equation}

The solution is defined in Equation~\ref{eq:aligned-algo}, where $\hat{G}$ is obtained by singular value decomposition (SVD). SVD decomposes $G$ into three matrices: $U$, $\Sigma$ and $V^\intercal$ where $U$ and $V$ are orthogonal matrices, and $\Sigma$ is a diagonal matrix containing the singular values of $G$. Algorithm of Aligned MTL is shown below: 

\begin{table}[H]
\centering
\resizebox{7.7cm}{!}{%
\begin{tabular}{l}
\toprule
Algorithm 2 Aligned MTL \\ \hline
\textbf{Require:} $G \in \mathbb{R}^{|\theta| \times T}$ – gradient matrix, \\
 \ \ \ \ \ \ \ \ \ \ \ \ \ \ \ \ $w \in \mathbb{R}^T $ – task importance  \\
\ \ $M \leftarrow G^\intercal G $ \\
\ \ $(\lambda, V) \leftarrow eigh(M)$ \\
\ \ $\Sigma^{-1} \leftarrow diag(\sqrt{\frac{1}{\lambda_1}},...,\sqrt{\frac{1}{\lambda_{R}}}) $ \\
\ \ $B \leftarrow \sqrt{\lambda_R}V \Sigma^{-1}V^\intercal $ \\
\ \ $\alpha \leftarrow Bw$ \\
\textbf{Return} $G\alpha$ \\
\bottomrule
\end{tabular}%
}
\label{tab:aligned_algo}
\end{table}

\end{document}